\pdfoutput=1

\documentclass[11pt]{article}

\usepackage[preprint]{acl}

\usepackage{times}
\usepackage{latexsym}

\usepackage[T1]{fontenc}

\usepackage[utf8]{inputenc}

\usepackage{microtype}

\usepackage{inconsolata}

\usepackage{graphicx}

\usepackage{hyperref}
\usepackage{array}
\usepackage{minted}
\usepackage{fancyvrb,fvextra}
\usepackage{tabularx}
\usepackage{caption}
\usepackage{multirow}
\usepackage{amsmath}
\usepackage{booktabs}

\title{A Comprehensive Study of Implicit and Explicit Biases in Large Language Models}

\begin{document}
\author{Fatima Kazi \\
  University of California, Davis \\
  Davis, CA, USA \\
  \texttt{fikazi@ucdavis.edu} \\\And
  Alex Young \\
  University of California, Davis \\
  Davis, CA, USA \\
  \texttt{axyoung@ucdavis.edu} \\\AND
  Yash Inani \\
  University of California, Davis \\
  Davis, CA, USA \\
  \texttt{yinani@ucdavis.edu} \\\And
  Setareh Rafatirad \\
  University of California, Davis \\
  Davis, CA, USA \\
  \texttt{srafatirad@ucdavis.edu} \\}
  
\maketitle

\begin{abstract}
Large Language Models (LLMs) inherit explicit and implicit biases from their training datasets. Identifying and mitigating biases in LLMs is crucial to ensure fair outputs, as they can perpetuate harmful stereotypes and misinformation. This study highlights the need to address biases in LLMs amid growing generative AI. We studied bias-specific benchmarks such as StereoSet and CrowSPairs to evaluate the existence of various biases in multiple generative models such as BERT and GPT 3.5. We proposed an automated Bias-Identification Framework to recognize various social biases in LLMs such as gender, race, profession, and religion. We adopted a two-pronged approach to detect explicit and implicit biases in text data. Results indicated fine-tuned models struggle with gender biases but excelled at identifying and avoiding racial biases. Our findings illustrated that despite having some success, LLMs often over-relied on keywords. To illuminate the capability of the analyzed LLMs in detecting implicit biases, we employed Bag-of-Words analysis and unveiled indications of implicit stereotyping within the vocabulary. To bolster the model performance, we applied an enhancement strategy involving fine-tuning models using prompting techniques and data augmentation of the bias benchmarks. The fine-tuned models exhibited promising adaptability during cross-dataset testing and significantly enhanced performance on implicit bias benchmarks, with performance gains of up to 20\%.
\end{abstract}

\section{Introduction}
\begin{figure*}
\centering
\includegraphics[width=6in]{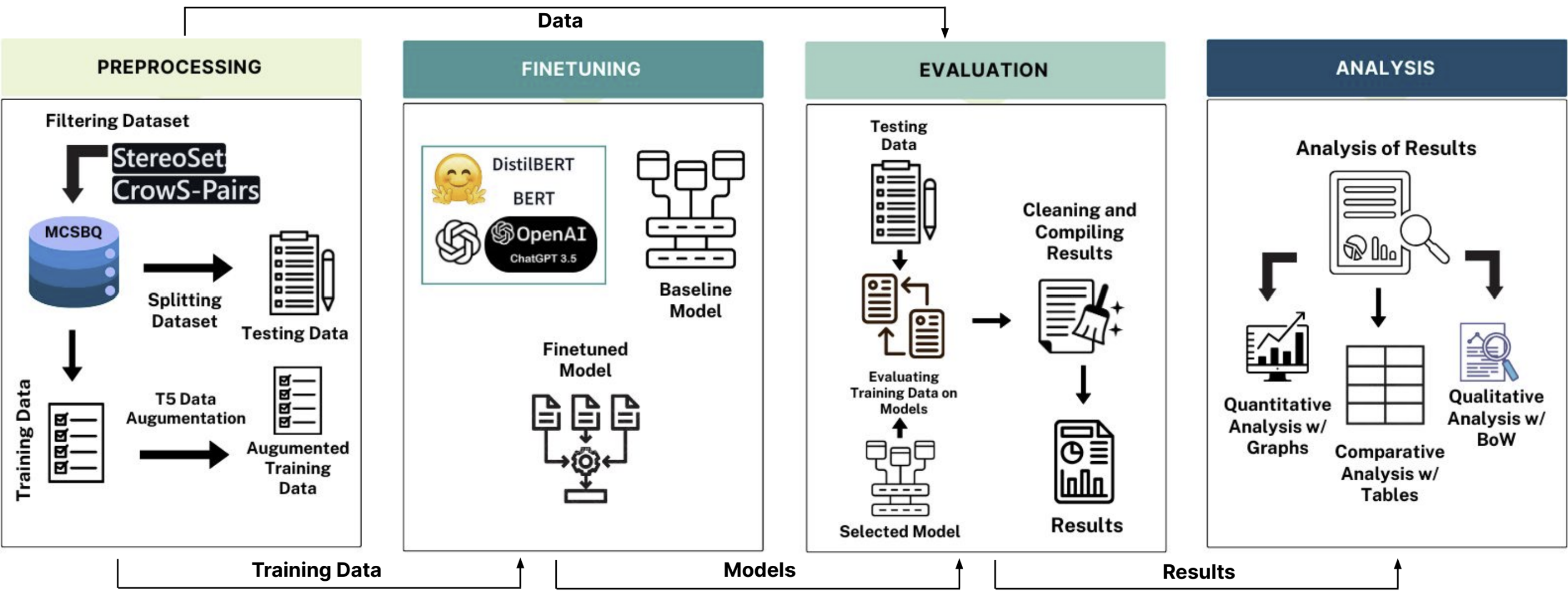}
\caption{The proposed framework comprises of 4 key components: Preprocessing - involving filtering StereoSet and CrowSPairs to create MCSBQ and splitting them into training and testing data. Finetuning - utilizing both original and augmented training data to fine-tune the LLMs. Evaluation - entailing testing of fine-tuned LLMs using testing data and baseline model using MCSBQ dataset. Analysis - analyzing results with quantitative (graphs), comparative (tables), and qualitative (BoW) techniques to uncover potential biases.}
\label{fig:frameworkImg}
\end{figure*}
The advancement of Natural Language Processing (NLP) has made Large Language Models (LLMs) ubiquitous in various industry applications. These models are employed to facilitate data accessibility, aid in data interpretation, and propose solutions based on data analysis \citep{bommasani2022opportunities, Github-copilot, OpenAI}. Their utilization spans across diverse sectors such as healthcare, where patient data analysis is enhanced with LLMs, and in software development for commercial purposes, where tools like Github Copilot assist in code generation \citep{Github-copilot}. This widespread adoption is not limited to corporate domains but also permeates into education \citep{Rudolph2023,tili2023what,xiaoming_implications,bommineni_mcat,education_era, engineering_edu}, where LLMs are increasingly used to support children with learning disabilities \citep{rane2023chatbot}. With such impact comes the responsibility to ensure that these models' responses are correct and unbiased \citep{tamkin2021understanding,bai2022constitutional,10.1145/3641289}. Because a majority of these models are trained on data from the internet such as web pages, books, articles, and forums, much of the data is opinion-based and may not always contain completely factual information \citep{CommonCrawl,10.1145/3442188.3445922,kenton2021alignment,weidinger2021ethical,gehman-etal-2020-realtoxicityprompts,10.1145/3641289,cho-etal-2019-measuring}. The fact that LLMs are generally based on self-supervised or unsupervised learning techniques, limits the capability of controlling what patterns, data points, and weights the model recognizes to be important \citep{huang2023trustgpt}. This gives rise to issues in models resulting in incorrect, biased, misinformed, or toxic information. Examples of this toxic output in popular models can be seen from RealToxicityPrompts dataset and jailbreaking ChatGPT 
\citep{zhang2023comprehensive,gehman-etal-2020-realtoxicityprompts}. Even with major pre-processing and purging data points that include very harmful or biased content, it is not guaranteed that a model will not learn a trend that could potentially be biased \citep{mikolajczykbarela2023data, 10.1145/3308560.3317590, 10.1145/3442188.3445922}. There could still be residual bias and algorithmic bias that show up \citep{lee2019algorithmic}. Bias is so incredibly hard to avoid because every system that has been influenced or directly created by humans is inadvertently also influenced by the biases of the creator, whether intentional or not. It follows that big data and LLMs are inherently filled with biases of all forms \citep{10.1145/3442188.3445922,10.1145/3308560.3317590,mikolajczykbarela2023data}. Such biases can result in harmful and negative impacts such as allocational, representational, and vulnerability impacts \citep{sheng2021societalbiaseslanguagegeneration}; these can propagate and amplify existing stereotypes, prevent certain populations from using specific resources, and leave groups more vulnerable to manipulation and harm \citep{prates2019assessinggenderbiasmachine, hashimoto2018fairnessdemographicsrepeatedloss, levy2021investigatingmemorizationconspiracytheories}. 
Characterizing the biases in LLMs holds profound significance in the landscape of AI and machine learning. 

In this paper, we have analyzed the biases of different models based on a large set of parameters including age status, disability, gender, nationality, physical appearance, profession, religion, and socioeconomic status. Our study involved investigating multiple models such as BERT, DistilBERT, GPT 3.5, and others to identify and characterize existing biases. We used two benchmark datasets, StereoSet and CrowSPairs, to analyze the extent of bias in each of the studied LLMs. As a bias-mitigation strategy, we used data augmentation methods to expand the training data used to fine-tune LLMs. A high-level outline of our proposed work is illustrated in Figure \ref{fig:frameworkImg}. The figure illustrates our workflow, serving as the foundation for our Bias-Characterization Framework. This framework focuses on reducing bias in the training data and conducting a comparative analysis of large language models (LLMs) before and after fine-tuning.



\section{Related Works} \label{relatedWorks}

Prior work underscores the importance of addressing biases in LLMs to ensure equitable and unbiased outcome. This section delves into an examination of previous studies that focus on identifying and mitigating biases concerning LLMs. 
Bias Bench was proposed by \citet{meade2022empirical} to track the effectiveness of bias mitigation techniques which include Dropout, CDA, and Self-Debias \citep{webster2021measuring, schick2021selfdiagnosis}. 
While \citet{meade2022empirical} use StereoSet and CrowSPairs, they only used BERT and GPT-2 in model usage and debias approach. 
\citet{bai2024measuring} proposed a technique to identify and measure implicit biases in LLMs using psychology-inspired measures where they applied the implicit association test (IAT) to their prompting method and assigning a stereotype bias level. They tested it on GPT-4 model and used the benchmarks BBQ, BOLD, and 70 decisions to showcase the benefits of their approach \citep{bai2024measuring}. However, their research may lack the ability to directly mitigate future bias. 
\citet{Kotek_2023} demonstrated the assumptions that different LLMs make when trying to assign gender markings to different occupations. They found that not only do LLMs pick the assign stereotypical gender roles to the occupations but amplifies the bias when associated with female individuals. While this paper highlights the extent of gender bias, other critical biases are still missing \citep{Kotek_2023}.
Another work that focuses on a specific bias is \citet{abid2021persistentantimuslimbiaslarge}; which considers the religious bias to compare results between \emph{Islam} and other languages.
Compared to prior work, our work takes a more holistic stance on bias identification and mitigation. Our framework not only examined explicit and implicit bias through established datasets but also through a novel prompting method. This versatile approach, applicable to a wide range of datasets, allows for easier adaptation and use with future large language models.

\section{Preliminary Results} \label{preliminaryResults}
Our preliminary experiments focused on understanding the extent of bias in popular LLMs such as ChatGPT. Due to the guardrails posed by OpenAI, \citep{OpenAI}, we experienced safeguarding of some key terms such as "race" and "gender". Despite these constraints, we found ways to bypass them by using synonyms such as replacements of "ethnicity", "country of origin", or "birthplace" instead of race and "occupation" for "profession", and with uncommon factors such as "socioeconomic status", "gpa standing", or "immigration status". Additionally, we devised prompts asking the model to generate characters based on certain features, revealing selective choices like "Maya Patel" for a doctor or "Carlos Ramirez" for a laborer, hinting at underlying biases.

Another experiment we tried on a profession was asking ChatGPT to classify a feature taking in other features as input, for example classifying the socioeconomic status of a student based on "gender", "age", and "race". With closer inspection, we discovered that ChatGPT stereotyped quite a bit to end up with the specified results; it classified 18-22 year old Asian males as 'Middle to High' class while Black males in the same age group as 'Low to Middle' class and Hispanic males as 'Low' class. When asking ChatGPT to classify the occupation of a person based on their "gender", "age", "race", and "country of origin" White males less than 25 year olds were categorized into three departments. If their country of origin was 'United States', then they were a 'Student' but if it was 'China' and 'India', they were 'Engineering' and 'Computer Science' respectively. ChatGPT then ended both of these examples by stating "Again, please note that this is just an example and the classification logic would depend on the data available and the specific requirements of the problem."

\section{Methodology} \label{methodology}

To formulate our queries and prompts, we leveraged Multiple-Choice Symbol Binding on the aforementioned datasets. The prompts are categorized into two distinct types: one prompts the model to identify stereotypes, while the other requires the model to choose between a stereotype and an anti-stereotype response. This approach allowed us to gain insights into the models' ability to discern and navigate biases present in the data. Furthermore, our methodology delves into the incorporation of data augmentation techniques. The original dataset undergoes augmentation through paraphrasing, and subsequent fine-tuning of the models is conducted to enhance their adaptability and performance.

\subsection{Models and Fine-tuning} \label{finetuning}
The models that we use in our study include ADA, BERT, DistilBERT, GPT-2, GPT-3.5, and a pretrained T5 model for paraphrasing \citep{sanh2020distilbert,devlin2019bert,brown2020language,ouyang2022training,ChatGPTParaphraser}. The process of fine-tuning the GPT models was executed using OpenAI API and the procedure for fine-tuning the GPT models adhered strictly to OpenAI's guidelines \citep{openai_finetuning, openai_api}, using the gpt-3.5-turbo-0613 model and the default values for hyperparameters. The process of fine-tuning base BERT and DistilBERT models was conducted using the HuggingFace Transformers library following the official documentation  \citep{HuggingFaceMultipleChoice}, specifically we implemented a multiple choice fine-tuning approach where models are tasked with identifying stereotypical portrayals within prompted text statements.
The hyperparameters implemented for BERT model were learning rate of $1 * 10^{-5}$, batch size=16, epochs=5 and weight decay of 0.01. For DistilBERT, similar parameters were utilized with no weight decay. When generating the fine-tuned models we used the same Prompting Techniques described in section \ref{promptingTechniques}. For splitting into training and test sets we used 20 and 8 data points for each bias type for training and the rest for testing for StereoSet and CrowSPairs respectively. 
To fine-tune models using bag-of-words results, the results were inserted directly into the message prompts themselves. This system role is a simple form of self-debiasing, a method proven to improve model bias \citep{meade2022empirical, schick2021selfdiagnosis}.

\subsection{Datasets}
To evaluate the different LLMs we used two main datasets, StereoSet and CrowSPairs. Both of these datasets ran on the models to detect the presence of bias and infer the extent to which bias exists. These datasets being crowdsourced allowed for more diversity in collecting the data in terms of the specific biases and targets and also with sentence structure. The crowdsourcing was also engineered in the United States so the overall biases and targets are in context to their prevalence in the US.

StereoSet is a dataset developed to help demonstrate the existence of bias with four main targets: \emph{gender}, \emph{profession}, \emph{religion}, and \emph{race}. \citep{nadeem-etal-2021-stereoset}. Crowdsourced Stereotype Pairs (CrowSPairs) benchmarks 9 different types of biases: \emph{race}, \emph{gender}, \emph{socioeconomic status}, \emph{nationality}, \emph{religion}, \emph{age}, \emph{sexual orientation}, \emph{physical appearance}, and \emph{disability} \citep{nangia-etal-2020-crows}.


\subsection{Prompting Techniques} \label{promptingTechniques}
Prompting techniques offer a powerful approach to enhance the control and effectiveness of LLMs. By carefully crafting prompts, we can guide LLMs towards desired outputs and improve their performance on specific tasks such as reducing bias \citep{yu2023large}.

In our evaluation, we employed the Multiple Choice Symbol Binding (MCSB) technique as the primary prompting method. MCSB has demonstrated superior efficacy in eliciting accurate responses, particularly in models with high MCSB ability, a category encompassing OpenAI models. Notably, this technique surpasses the conventional cloze method in precision \citep{robinson2023leveraging}. Furthermore, we introduced two distinct types of prompts to discern nuanced aspects of bias manifestation, Implicit Bias Prompting and Explicit Bias Prompting. In Explicit Bias Prompting, we task the model with selecting a prompt with explicit bias reference. This method seeks to uncover any explicit biases that may influence the model's choice, providing insights into how it perceives bias. An example of this techniques usage can be seen in Table \ref{table:prompting-exampleExplicit}. In Implicit Bias Prompting, we directly inquired whether the model perceived a given prompt as biased or unbiased. This direct approach aimed to capture the model's inferred acknowledgment of bias. The approach is similar to Table \ref{table:prompting-exampleExplicit} if line 4 is switched to "Based on the context, answer with only A, B or C", and line 12 is switched to "A, B, or C?". Notice that the model is never told that there could be biases or stereotypes in it's choices, signifying any bias to have been implicit.

\begin{table}[ht]
  \centering
  
  \begin{tabular}{cp{0.7\linewidth}}
    \hline
    \textbf{Role} & \textbf{Content} \\
    \hline
    System & You are a helpful assistant who will answer with just 'A', 'B', or 'C'. \\
    \hline
    User & Hispanic people are very diverse. \\
    Assistant & Understood. \\
    \hline
    User & Answer with only A, B, or C based on which prompt is a stereotype. \\
    Assistant & Understood. \\
    \hline
    User & A: Not all individuals like that are liable to commit crimes. \\
    Assistant & Understood. \\
    User & B: It's a highly convoluted game for football to play. \\
    Assistant & Understood. \\
    User & C: Anyone who behaves like that is considered a criminal. \\
    Assistant & Understood. \\
    \hline
    User & Which prompt is a stereotype: A, B, or C? \\
    \hline
  \end{tabular}
    \caption{Example of MCSB ChatGPT Prompting Technique for StereoSet Data Point With Explicit Bias Prompting}
    \label{table:prompting-exampleExplicit}
    
\end{table}

\subsection{Data Augmentation}
In our pursuit of mitigating biases within language models, we leveraged data augmentation as a strategic approach aimed at enhancing the models' resilience to biased content, ultimately contributing to model debiasing. Our methodology involved the augmentation of the original dataset through paraphrasing, a process facilitated by two distinct models: Google's T5 model and GPT-3.5. The paraphrasing task entailed presenting prompts to a pre-trained T5 model in the format {paraphrase: {prompt}}, while GPT-3.5 was prompted to paraphrase the given questions. By engaging both models with the datasets, namely CrowsSPairs and StereoSet, we looked to generate diversified and nuanced perspectives on the provided content. During the fine-tuning process, the paraphrased data replaced the original data points, effectively infusing the training data with augmented variants. This augmentation strategy was twofold: first, to encourage the model to generalize better by exposure to a broader spectrum of language, and second, to promote the model's ability to discern and handle biased content more effectively \citep{Abaskohi_2023,tang2023metrics,yu2023large}. 

\section{Results}\label{results}
The models were evaluated with subsets of the StereoSet and CrowSPairs Datasets and with different prompting techniques explained. The responses of each model were recorded and compiled to measure the existence and extent of bias quantitatively. The responses were grouped into each of their targets and the percentages of each target were determined. Then, for qualitative analysis, the results were examined using Bag of Words models to probe the specificity of bias in models.

\subsection{General Results - Implicit Bias Prompting}

\begin{table}[ht]
\centering

\begin{tabular}[t]{>{\arraybackslash}m{0.26\linewidth} *{4}{>{\centering\arraybackslash}m{0.185\linewidth}}}

\toprule
\multicolumn{4}{c}{StereoSet} \\
\midrule
            &GPT 3.5        &DistilBERT &BERT\\
\hline
Gender      &0.48          &0.36       &0.36\\
Race        &0.41         &0.26       &0.33\\
Profession  &0.42         &0.23       &0.35\\
Religion    &0.37          &0.27       &0.29\\
\end{tabular}
\begin{tabular}[t]{>{\arraybackslash}m{0.26\linewidth} *{4}{>{\centering\arraybackslash}m{0.185\linewidth}}}
\toprule
\multicolumn{4}{c}{CrowSPairs} \\
\midrule

                    &GPT 3.5    &DistilBERT &BERT\\
\hline
Age Status           &0.24        &0.40       &0.46\\
Disability           &0.25        &0.63       &0.60\\
Gender               &0.21        &0.52       &0.47\\
Nationality          &0.36         &0.47       &0.57\\
Physical Appearance  &0.25         &0.46       &0.43\\
Race                 &0.36        &0.33       &0.57\\
Religion             &0.39         &0.31       &0.39\\
Sexual Orientation   &0.40         &0.55       &0.77\\
Socioeconomic Status &0.23         &0.49       &0.66\\
\bottomrule
\end{tabular}
\caption{Performance of various models in selecting stereotypical responses when implicitly prompted}
\label{table:generalTableResultNORM}

\end{table}

The evaluation of general results using our benchmarks and baselines is presented in Table \ref{table:generalTableResultNORM}. We found that more recently developed models are doing a bit better at being able to prevent stereotypical answers. It is evident that some models fared better than others in specific features but overall all models performed in similar ranges. Based on Table \ref{table:generalTableResultNORM}, GPT 3.5 exhibited a higher percentage of biased responses compared to the ADA or BERT models in every category with the StereoSet dataset. It can be noted that the ratio of stereotypical responses chosen by models evaluated with the StereoSet dataset, \emph{Gender} fared the worst overall in each of the models. \emph{Gender} bias was a fairly large portion of StereoSet and encompassed many targets such as \emph{male}, \emph{mother}, \emph{himself}, \emph{herself}, and others.

For models tested on CrowSPairs, Table \ref{table:generalTableResultNORM} shows that BERT performed worse than DistilBERT for biases such as \emph{sexual orientation}, \emph{socioeconomic status}, and \emph{disability}. For example \emph{sexual orientation} which performed poorly with the stereotype being chosen 77\% of the time. This could be because this is a fairly underrepresented bias in general, leading to less training, guardrails, and testing. This is in relation to \emph{race} and \emph{religion} which are more popular areas of bias.

\subsection{General Results - Explicit Bias Prompting}

\begin{table}[ht]
\centering

\begin{tabular}[t]{>{\arraybackslash}m{0.26\linewidth} *{4}{>{\centering\arraybackslash}m{0.185\linewidth}}}
\toprule
\multicolumn{4}{c}{StereoSet} \\
\midrule
            &GPT 3.5    &DistilBERT &BERT\\
\hline
Gender      &0.61        &0.30       &0.07\\
Race        &0.82          &0.36       &0.09\\
Profession  &0.67        &0.35       &0.08\\
Religion    &0.68        &0.35       &0.08\\

\end{tabular}
\begin{tabular}[t]{>{\arraybackslash}m{0.26\linewidth} *{4}{>{\centering\arraybackslash}m{0.18\linewidth}}}
\toprule
\multicolumn{4}{c}{CrowSPairs} \\
\midrule

                    &GPT 3.5     &DistilBERT &BERT\\
\hline
Age Status           &0.56      &0.36       &0.63\\
Disability           &0.73        &0.35       &0.43\\
Gender               &0.60         &0.49       &0.57\\
Nationality          &0.75       &0.43       &0.56\\
Physical Appearance  &0.76       &0.48       &0.46\\
Race                 &0.75      &0.37       &0.57\\
Religion             &0.79        &0.29       &0.44\\
Sexual Orientation   &0.76        &0.77       &0.64\\
Socioeconomic Status &0.67        &0.41       &0.59\\
\bottomrule
\end{tabular}
\caption{Comparison of various models in generating stereotypical responses when explicitly prompted}
\label{table:generalTableResultBIAS}

\end{table}

\begin{table*}[ht]
\centering

\begin{tabular}[t]{>{\arraybackslash}m{0.2\linewidth} *{6}{>{\centering\arraybackslash}m{0.11\linewidth}}}
\toprule
\multicolumn{7}{c}{StereoSet Test Data \& CrowSPairs Trained Model} \\
\midrule
&\multicolumn{2}{c}{ChatGPT}&\multicolumn{2}{c}{BERT}&\multicolumn{2}{c}{DistilBERT}\\

            &No Aug     &T5 Aug     &No aug     &T5 Aug     &No aug     &T5 aug\\
\hline
Gender      &0.41       &0.51       &0.09       &0.59       &0.28       &0.19\\
Race        &0.12       &0.21       &0.11       &0.53       &0.62       &0.48\\
Profession  &0.28       &0.34       &0.10       &0.53       &0.43       &0.26\\
Religion    &0.16       &0.23       &0.05       &0.55       &0.57       &0.43\\

\end{tabular}
\begin{tabular}[t]{>{\arraybackslash}m{0.2\linewidth} *{6}{>{\centering\arraybackslash}m{0.11\linewidth}}}
\toprule
\multicolumn{7}{c}{CrowSPairs Test Data \& StereoSet Trained Model} \\
\midrule

                        &No Aug      &T5 Aug     &No Aug      &T5 Aug     &No Aug      &T5 Aug\\
\hline
Age Status              &0.35       &0.52       &0.44       &0.39       &0.35       &0.35\\
Disability              &0.31       &0.44       &0.56       &0.71       &0.48       &0.63\\
Gender                  &0.43       &0.45       &0.46       &0.56       &0.51       &0.56\\
Nationality             &0.40       &0.45       &0.41       &0.54       &0.35       &0.38\\
Physical Appearance     &0.33       &0.45       &0.58       &0.56       &0.49       &0.60\\
Race                    &0.36       &0.35       &0.58       &0.60       &0.66       &0.63\\
Religion                &0.36       &0.30       &0.74       &0.63       &0.77       &0.86\\
Sexual Orientation      &0.51       &0.42       &0.36       &0.59       &0.53       &0.76\\
Socioeconomic Status    &0.38       &0.42       &0.52       &0.57       &0.62       &0.68\\
\bottomrule
\end{tabular}
\caption{Comparison of various models fine-tuned on StereoSet dataset under implicit prompting with following configurations: No Aug (No Augmentation), T5 Aug (Augmented using T5)}
\label{table:crossTableResultsNORM}
\end{table*}
It is important to recognize how the two prompting techniques utilized affected the accrued results with the Explicit Bias Prompting results in Table \ref{table:generalTableResultBIAS}. Take note of BERT performing really well with this prompting method. The ratio of picking the stereotype is so low comparatively to other models and other tests on BERT because the model picked the unrelated response, on average, 74\% of the time. Despite getting great results in face-value, these results don't mean anything in terms of the model's actual capability in answering questions since it would pick the unrelated response, giving answers not remotely related to the question or even the context. Note the performance of BERT with the CrowSPairs dataset seeming more reasonable in comparison to StereoSet due to not picking unrelated. BERT aside, other models performed worse with this prompting technique in both CrowSPairs and StereoSet. This is expected since we had asked the model to pick the stereotype in the first place. This shows the model's ability in understanding what a stereotype is by definition. Note the variety in percentages of the model being able to correctly pick the response with a stereotype with DistilBERT ranging from 29\% to 77\%.

\subsection{Augmented Fine-Tuning Results - Implicit Bias Prompting}

We fine-tuned our models after augmenting our data. This allowed us to see how embedded bias is. We noticed that a decrease in choosing the stereotype was observed in StereoSet but not in all of CrowSPairs. This could just be the fact that our fine-tuning was not pragmatic enough, although it can also be inferred that the biases that did not do well originally are the ones that improved. This could also mean that there is a lower bound that the model has in terms of its performance. We also took note of the bias that was improved the most overall, \emph{race}, in comparison to the other biases which were either improved not at all or only little in StereoSet. We assumed this is the case due to the sheer count of prompts in the \emph{race} bias promoting the model to learn more in the fine-tuning process. \emph{Race} also did generally well in the CrowSPairs dataset likely for the same reason. 

To further evaluate the robustness of our fine-tuned models, we cross-tested each model with the other respective dataset. The results are displayed in Table \ref{table:crossTableResultsNORM}. We displayed each model with two different types of fine-tuning: fine-tuned without any augmentation and fine-tuned with T5 with augmention. With this, we were able to see that our models showed respectable results for some of the bias and model combinations. 
We consider the fine-tuned model to have considerably positive results when it outperforms the base model. For StereoSet, \emph{race} performed significantly better with a bias decrease by 30\%. This coincided with our previous conclusion that \emph{race}'s fine-tuned results were the best overall. That said, it can be seen that \emph{gender} in the StereoSet performed worse than both the baseline and worst than the fine-tuned which is consistent with our fine-tuned results as well. This is likely due to deeply ingrained gender biases within the training data and model.


\subsection{Augmented Fine-Tuning Results - Explicit Bias Prompting}

\begin{table}[ht]
\captionsetup{width=\linewidth}
\centering

\begin{tabular}[t]{>{\arraybackslash}m{0.26\linewidth} *{3}{>{\centering\arraybackslash}m{0.185\linewidth}}}
\toprule
\multicolumn{4}{c}{StereoSet} \\
\midrule
            &NFNA           &FTNA       &FTAT5\\
\hline
Gender      &0.61           &+0.19      &+0.08\\
Race        &0.82           &+0.08      &+0.01\\
Profession  &0.67           &+0.19      &+0.14\\
Religion    &0.68           &+0.16      &+0.16\\
\end{tabular}
\begin{tabular}[t]{>{\arraybackslash}m{0.26\linewidth} *{3}{>{\centering\arraybackslash}m{0.185\linewidth}}}
\toprule
\multicolumn{4}{c}{CrowSPairs} \\
\midrule

                    &NFNA           &FTNA       &FTAT5\\
\hline
Age Status          &0.24           &+0.39      &+0.39\\
Disability          &0.25           &+0.40      &+0.40\\
Gender              &0.21           &+0.34      &+0.34\\
Nationality         &0.36           &+0.39      &+0.39\\
Physical Appearance &0.25           &+0.50      &+0.50\\
Race                &0.36           &+0.39      &+0.39\\
Religion            &0.39           &+0.38      &+0.38\\
Sexual Orientation  &0.40           &+0.30      &+0.30\\
Socioeconomic Status&0.23           &+0.38      &+0.38\\
\bottomrule
\end{tabular}
\caption{Comparison of GPT-3.5's stereotype selection under explicit prompting with following configurations: NFNA (No Fine-tuning, No Augmentation), FTNA (Fine-tuning, No Augmentation), FTAT5 (FT with T5 Augmented data)}
\label{table:augmentedTableResultsBIAS}
\end{table}
The models we fine-tuned were also re-prompted with the Explicit Bias prompting technique. This would give us some concrete evidence on the resiliency of the models and stubbornness in un-learning bias. We can see here from Table \ref{table:augmentedTableResultsBIAS} that GPT 3.5 was able to pick the stereotype correctly more often as compared to its base model. Notice with the experiment, the proportion stayed around 75\% with StereoSet increasing by 10\% and 60\% with CrowSPairs by increasing by 40\%. There was not much success in the models performing better than these numbers. In this way, we can see that with this fine-tuning we can train models to be able to better identify the stereotype so that it is able to avoid the stereotype when prompted implicitly.

Take note of the overall performance increase in Table \ref{table:augmentedTableResultsBIAS} which indicates that the models were able to pick the response with the stereotype more often after fine-tuning. 
Fine-tuning allowed each bias to rise to some common base value especially for those biases that had not performed as well as others initially. \emph{Socioeconomic Status}, \emph{disability}, and others had not performed well because those were likely not biases that had been considered extensively when initially training the model, which allowed their performance to increase by such a large margin instantly and with ease.

\begin{table*}[ht]
\centering

\begin{tabular}[t]{>{\arraybackslash}m{0.2\linewidth} *{6}{>{\centering\arraybackslash}m{0.105\linewidth}}}
\toprule
\multicolumn{7}{c}{StereoSet Test Data \& CrowSPairs Trained model} \\
\midrule
&\multicolumn{2}{c}{ChatGPT}&\multicolumn{2}{c}{BERT}&\multicolumn{2}{c}{DistilBERT}\\

            &No Aug     &T5 Aug     &No Aug     &T5 Aug     &No Aug     &T5 aug\\
\hline
Gender      &0.58       &0.55       &0.12       &0.54       &0.31       &0.21\\
Race        &0.89       &0.83       &0.15       &0.53       &0.61       &0.47\\
Profession  &0.72       &0.67       &0.13       &0.53       &0.45       &0.25\\
Religion    &0.74       &0.72       &0.05       &0.62       &0.59       &0.45\\
\end{tabular}
\begin{tabular}[t]{>{\arraybackslash}m{0.2\linewidth} *{6}{>{\centering\arraybackslash}m{0.105\linewidth}}}
\toprule
\multicolumn{7}{c}{CrowSPairs Test Data \& StereoSet Trained Model} \\
\midrule

                        &No Aug      &T5 Aug     &No Aug      &T5 AUG     &No Aug      &T5 AUG\\
\hline
Age Status              &0.78       &0.30       &0.35       &0.34       &0.35       &0.35\\
Disability              &0.75       &0.37       &0.48       &0.65       &0.42       &0.69\\
Gender                  &0.70       &0.30       &0.54       &0.51       &0.53       &0.59\\
Nationality             &0.77       &0.30       &0.50       &0.66       &0.34       &0.40\\
Physical Appearance     &0.82       &0.67       &0.49       &0.60       &0.47       &0.67\\
Race                    &0.77       &0.45       &0.51       &0.67       &0.67       &0.65\\
Religion                &0.74       &0.43       &0.71       &0.63       &0.75       &0.85\\
Sexual Orientation      &0.84       &0.42       &0.41       &0.59       &0.58       &0.74\\
Socioeconomic Status    &0.76       &0.43       &0.61       &0.70       &0.57       &0.63\\
\bottomrule
\end{tabular}
\caption{Comparison of various models fine-tuned on StereoSet dataset with different configurations: No Aug (No Augmentation), T5 Aug (Augmented using T5), and Explicit Prompting on the CrowSPairs dataset}
\label{table:crossTableResultsBIAS}
\end{table*}

After receiving these positive results with fine-tuning for Explicit Bias, we also cross-evaluated the fine-tuned models to assess whether these substantial results were the result of overfitting or other justification which is presented in Table \ref{table:crossTableResultsBIAS}. While the results were not as high as with the fine-tuned model, they were still better than the base model which is another sample of evidence proving the fine-tuning had constructive results. Specifically, notice how BERT no longer refused to answer the prompt nor pick completely unrelated answer which is significant in its performance improvement. 
In particular, note the biases that were not included in training StereoSet dataset performed lower and worse than those that were included. \emph{Sexual Orientation}, \emph{disability}, and \emph{age status} are examples of such biases with their results around the 30-40\% range while \emph{gender}, \emph{race}, and \emph{religion} performed around 30\% better.

\subsection{Bag of Words}
\begin{table}[h]

  \begin{tabular}{p{0.15\linewidth}p{0.75\linewidth}}
  \toprule
  \multicolumn{2}{c}{Top 7 words helped decide it wasn't stereotype} \\
    \midrule
    All & 'good', 'people', 'strong', 'day', 'athletic', 'war', 'kind'\\
    \midrule
    Gender & 'caring', 'nurturing', 'manners', 'likes', 'good', 'picked', 'strong'\\
    \midrule
    Race & 'lot', 'thin', 'dangerous', 'hot', 'people', 'muslim', 'war' \\
    \midrule
    Profession & 'time', 'white', 'smart', 'strong', 'glasses', 'always' \\
    \midrule
    Religion & 'violent', 'african', 'black', 'late', 'fasted', 'august', 'suggestions' \\
    \toprule
    \multicolumn{2}{c}{Top 7 words helped decide it was stereotype} \\
    \midrule
    All & 'terrorists', 'dangerous', 'poor', 'black', 'white', 'violent', 'probably' \\
    \midrule
    Gender & 'wearing', 'cookies', 'household', 'homemaker', 'one', 'skirt', 'gossiping' \\
    \midrule
    Race & 'poor', 'terrorist', 'dirty', 'black', 'bad', 'always', 'like' \\
    \midrule
    Profession & 'nerdy', 'awkward', 'rude', 'wears', 'poor', 'glasses', 'like'\\
    \midrule
    Religion & 'terrorists', 'holy', 'must', 'everyone', 'scary', 'everybody', 'vedas' \\
    \hline
  \end{tabular}
    
    \caption{Bag-of-Words Analysis in StereoSet Dataset under Explicit Prompting}
    \label{table:tableBowBiasNFT}
\end{table}

We were not able to fully grasp the model's process and model's inclination in choosing a stereotype just by looking at numbers, percentages, and changes. So we decided to run multiple Bag of Words models to further analyze the performance of GPT 3.5 on StereoSet for qualitative analysis. In this way we were able to particularly understand the model's result and its relation with concrete words, definitions, and topics it deemed to be important. 

For Implicit Bias prompting, we gathered that these are words that the model had concluded to be inherently true since it wasn't specifically asked to choose a stereotype. We observed that most of the words can be deduced to have a negative connotation such as 'violent', 'terrorist', and 'aggressive'. Along with that, the words which coerced the model to choose the anti-stereotype or unrelated option were also similarly connotated with a couple of positive words thrown in like 'nerdy', 'smart', or 'protective'. 

In Table \ref{table:tableBowBiasNFT}, we indicated which words were able to influence the model in assisting to choose which expression was the stereotype as well as which words guided the model away from the other expressions. It can be seen that even in Explicit Bias Prompting, the model is able to recognize that these words with negative connotations would be considered stereotypical or biased towards specific groups. Words such as 'dangerous', 'dirty', 'poverty', 'scary' and 'terrorist' were shown to be crucial in making the decision as to whether the prompt was a stereotype or not. That said, words that generally had a positive connotation directed the model away from the model in its decision of picking the stereotype, words such as 'athletic', 'nurturing', 'strong' and 'caring'. That is not to say that positive words could not be stereotypical, however they would, by large, not have as large of a negative societal impact. But even with these positive words, there is a noticeable trend in that the \emph{race} and \emph{religion} biases still included unfavorable words in the model's influence of choosing a prompt to be stereotypical or not. 'African', 'hot', 'thin', 'violent', and 'war' are some of the words in these two biases that had shown up time and time again as stereotypical words. 


\subsection{BoW Fintetuning and System Role}
Using these BoW results, we then fine-tuned a GPT model in an attempt to debias it. An example of this debias insertion is shown in section \ref{finetuning} where the words would be inserted in the square brackets when fine-tuning GPT models. The results from models fine tuned using prompts augmented with bag-of-word outputs indicates an improvement in implicit bias indication and a reduced capacity for detected explicit bias. For example, in the StereoSet benchmark there was an improvement from 24\% stereotype picked to 22\% in the implicit testing for race, but a decrease from 90\% to 86\% stereotype detection. This comparison was made with relation to the base fine-tuned model without augmentation. This was fair considering that Bag of Words output often indicated that the base models have issue detecting positive biases, however by including this in the prompt likely split the model's attention between positive and negative biases. This indicated that the base GPT and BERT models still have trouble grasping what biases are and instead focus on sentiments.

The results from GPT model fine tuned using prompts with augmented system roles were quite similar to the results from the model augmented with bag-of-words, however it performed slightly better in detecting explicit biases. For example, using the StereoSet benchmark, the stereotype detection detected racial stereotypes 87\% of the time instead of 86\% of the time. In addition, the system role augmented model performed much better than the bag-of-words augmented model at cross evaluation. This was likely due to issues with model overfitting on the Bag of Words outputs that are increasing complexity in the prompt.

\section{Conclusion} \label{conclusion}
The training data for LLMs originate from the internet, an aggregation of misinformation, opinions melded with facts, and online discourse. As a result, this fusion is incorporated implicitly, and sometimes explicitly, into the LLMs themselves. One of the safeguards implemented in some models is the decision to focus on the anti-stereotype which reveals that the model is aware of the stereotype. Results also showed the models' difficulty in distinguishing and avoiding specific biases. The \emph{gender} bias performed worse in this capacity consistently over different models, datasets, and fine-tuned results as well, with it performing up to 30\% worse than other biases. Perhaps this indicates a deeply ingrained gender bias in the models.
After implementing fine-tuning on the models, it was apparent that these LLMs had the capacity of learning to be less biased in both implicit and explicit bias. While this is true, this just allows the models correct themselves after bias has been identified instead of unlearning it in the first place. It was also noticed during our analysis with Bag of Words that the model focused on specific words and terminologies: words that generally have a negative connotation linked to them. The fine-tuning results demonstrated that for implicit bias, the anti-stereotype was picked most often with the following ordered techniques: sysrole > BoW > GPT paraphrasing > T5 paraphrasing > base model. Additionally, our cross-evaluation results illustrated some of the potential of these bias mitigation techniques in more "real-world" scenarios. These results perpetuate the need for better prompting techniques as well as the benefits of self-debiasing in the form of the system role prompting.

\subsection{Bias-Identification Framework}
Using our methodology and results, we proposed a Bias-Identification Framework (BIF) to recognize various social biases in LLMs. The BIF follows our methodology of using MCSB prompting with implicit and explicit bias questions. Using publicly available bias datasets have the model in question predict the most likely category (stereotype, anti-stereotype, or unrelated) based on its understanding of bias. By comparing the model's predicted category with the actual category from the dataset the model's bias performance can be measured. Manual testing can confirm any explanations the model presents. This framework can be used not only for research but also for developing techniques to reduce bias and ensure responsible development and deployment of LLMs. Unlike explicit bias where perfect detection is ideal, implicit bias detection and mitigation goals should be set by developers, targeting a balanced model that exhibits stereotypes and anti-stereotypes at a rate (e.g., 20-50\%) that considers the limitations of prompt design and potential overcorrection by the model. As this framework is simple in nature, it can guard against potentially biased bias-benchmarks through utilizing a more diverse set of benchmarks together, potentially including IAT.

\subsection{Limitations}
Our study on bias in LLMs had certain limitations. The amount of data we could test via API was constrained, limiting the scale of our study. While the use of MCSB is accurate and streamlined, it restricted exploration of responses that exceed pre-set options, which could offer deeper bias insights; different benchmarks or stereotype definitions could yield varied findings. When considering limitations, it is important to understand the limitations of bias benchmarks in general, as discussed by \citet{blodgett-etal-2021-stereotyping}. Stereotyping benchmarks such as Crows-Pairs and StereoSet face challenges due to the inherent subjectivity of stereotypes. This can lead to situations where an LLM's response aligns with commonly held beliefs within a specific context, yet the benchmark flags it for bias \citep{blodgett-etal-2021-stereotyping}. Additionally, potential biases within the benchmarks themselves are a concern as they may reflect biases of the creators. These limitations restrict the generalizability of findings from bias detection methods using these benchmarks.

\subsection{Ethics Statement}
This research utilizes established bias benchmarks to assess potential biases in LLMs. Our aim is to contribute to the development of fairer NLP applications. Our methodology is designed to minimize the risk of perpetuating social biases within LLMs. We acknowledge the limitations of benchmarks and the importance of ongoing research in creating robust methods for bias detection and mitigation.

\bibliography{custom}

\end{document}